\documentclass[preprint]{iris-ai-25} 

\usepackage{inconsolata}
\usepackage{todonotes}
\usepackage{rotating}
\usepackage{graphicx}
\usepackage{balance} 
\usepackage{lscape}
\usepackage{amsmath}
\usepackage{amssymb}
\usepackage{amsthm}
\usepackage{float}
\usepackage{multirow}
\usepackage{algorithm}
\usepackage{algpseudocode}
\usepackage{tcolorbox}
\usepackage{booktabs}
\usepackage{pifont}
\usepackage{bbm}
\usepackage[normalem]{ulem}
\usepackage{hyperref}
\usepackage{algorithmicx}

\title{Evaluating Knowledge Graph Based Retrieval Augmented Generation Methods under Knowledge Incompleteness}

\author{%
Dongzhuoran Zhou\textsuperscript{1,2},
Yuqicheng Zhu\textsuperscript{2,3}, 
Xiaxia Wang\textsuperscript{4},
Hongkuan Zhou\textsuperscript{2,3}, \\
\textbf{Yuan He\textsuperscript{4}},
\textbf{Jiaoyan Chen\textsuperscript{5}},
\textbf{Steffen Staab\textsuperscript{3,7},}
\textbf{Evgeny Kharlamov\textsuperscript{1,2}}
\\
\textsuperscript{1}University of Oslo, 
\textsuperscript{2}Bosch Center for AI, 
\textsuperscript{3}University of Stuttgart, 
\\
\textsuperscript{4}University of Oxford, 
\textsuperscript{5}The University of Manchester, 
\textsuperscript{7}University of Southampton\\
\texttt{dongzhuoran.zhou@de.bosch.com}\\
}

\begin{document}

\maketitle

\begin{abstract}
Knowledge Graph based Retrieval-Augmented Generation (KG-RAG) is a technique that enhances Large Language Model (LLM) inference in tasks like Question Answering (QA) by retrieving relevant information from knowledge graphs (KGs).
However, KGs are often incomplete, meaning that essential information for answering questions may be missing. Existing benchmarks do not adequately capture the impact of KG incompleteness on KG-RAG performance. In this paper, we systematically evaluate KG-RAG methods under incomplete KGs by removing triples using different methods and analyzing the resulting effects. We demonstrate that KG-RAG methods are sensitive to KG incompleteness, highlighting the need for more robust approaches in realistic settings.
\end{abstract}

\section{Introduction}
Large Language Models (LLMs) have achieved remarkable success across various natural language processing tasks such as Question Answering (QA) \citep{achiam2023gpt, touvron2023llama, jiang2023mistral}.
However, they face several limitations, including outdated knowledge \citep{dhingra2022time}, insufficient domain-specific expertise \citep{li2023chatgpt}, and a tendency to generate plausible but incorrect information, known as hallucination \citep{ji2023survey}.

Retrieval-Augmented Generation (RAG) mitigates these issues by integrating information retrieval mechanisms, allowing LLMs to access up-to-date and reliable information without requiring modifications to their architecture or parameters \citep{khandelwal2019generalization, ram2023context, borgeaud2022improving, izacard2021leveraging}. While conventional RAG approaches rely on unstructured text corpora, recent studies \citep{chen2023contextual, han2024retrieval, zhang2025survey} have started to explore knowledge graphs (KGs) as knowledge resources by exploring different retrieval methods, such as neighborhood extraction after entity linking and similarity-based entity search in the embedding space, as well as different prompting methods like knowledge re-writting \citep{wu2024cotkr}.
Such KG-based RAG (KG-RAG for shortness) methods can reduce text redundancy \citep{li2024simple}, enable flexible updates \citep{paulheim2016knowledge}, and provide structured reasoning evidence \citep{luo2024reasoning}. Despite these benefits, KGs are often incomplete in real-world applications \citep{min2013distant, renquery2box, pflueger2022gnnq}, raising a crucial question: \emph{Can KG-RAG methods remain effective when the given KG is incomplete}?

LLMs have demonstrated substantial knowledge retention and reasoning capabilities \citep{achiam2023gpt,guo2025deepseek}. Consequently, KG-RAG methods \citep{luo2024reasoning, sun2024think, he2024g} are expected to leverage LLMs to infer relevant facts that extend beyond explicitly retrieved triples. For instance, from the retrieved triples \(\langle \textit{JustinBieber, has\_parent, JeremyBieber} \rangle\) and \(\langle \textit{JeremyBieber, has\_child, JaxonBieber} \rangle\), LLMs can infer that Jaxon Bieber is Justin Bieber’s brother by recognizing the underlying semantic path. In this way, KG-RAG methods can effectively compensate for missing direct evidence and mitigate the impact of incomplete KGs.

However, existing KG-RAG studies directly adopt KGQA benchmarks \citep{yih2016value,talmor2018web} for evaluation.
In these benchmarks, the ground truth answer of each question can be directly inferred from the knowledge in the KG, 
failing to reflect real-world scenarios where KGs are incomplete for answering the question. 
Thus the current evaluation does not reflect KG-RAG methods' performance facing incomplete knowledge.
In this paper, we systematically assess three popular KG-RAG methods under varying levels of KG incompleteness. Our findings reveal that \textbf{these methods still benefit significantly from incomplete KGs, but are sensitive to missing knowledge with performance drops}, emphasizing the need for more robust approaches for KG-RAG.

\begin{figure}[t] 
    \centering
    \includegraphics[width=0.65\textwidth]{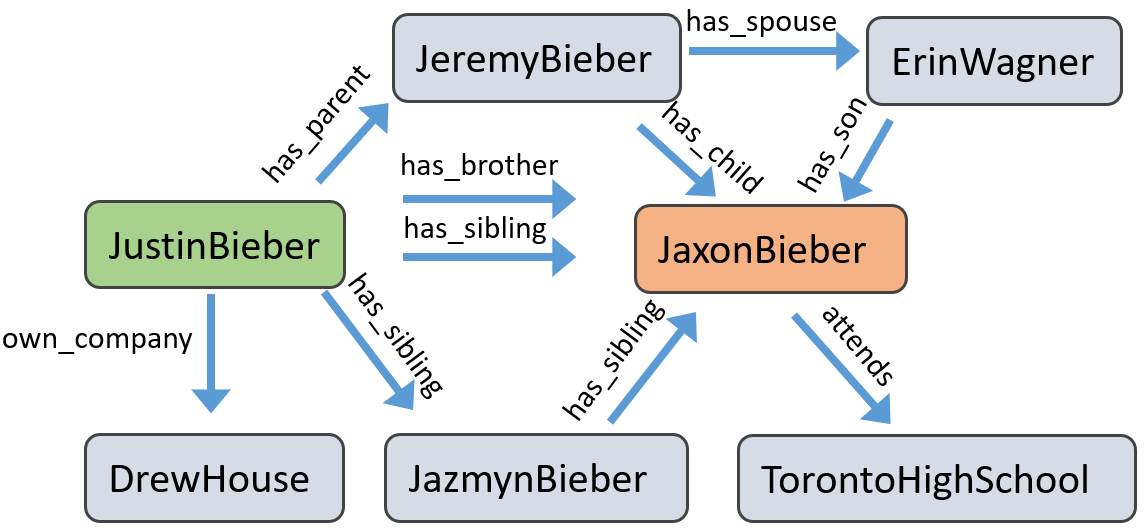} 
    \caption{\textbf{Illustration of the running example.} The question ``\textit{Who is the brother of Justin Bieber?}'' has the expected answer \textit{JaxonBieber}. While the direct evidence path \textcolor{blue}{$\textit{JustinBieber} \xrightarrow{\textit{has\_brother}} \textit{JaxonBieber}$} provides a straightforward answer, it is often missing in practice.
    A robust KG-RAG method should leverage alternative reasoning paths, such as \textcolor{blue}{$\textit{JustinBieber} \xrightarrow{\textit{has\_parent}} \textit{JeremyBieber} \xrightarrow{\textit{has\_child}} \textit{JaxonBieber}$}, to derive the correct answer.}
    \label{fig:example}
\end{figure}

\section{Methodology}
\subsection{Background and Notations}

\paragraph{Knowledge Graph.} Given two finite sets $E$ and $R$, whose elements are called \emph{entities} and \emph{relations}, a KG $G$ is a directed multi-relational graph represented as a subset of $E\times R\times E$, where each element $\langle h,r,t\rangle\in G$ is called a \emph{triple}.

\paragraph{Reasoning Path.} A reasoning path in a KG is a directed sequence of entities and relations that connect a source entity to a target entity, forming a logical pathway for inference:
\begin{equation*}
    e_{0} \xrightarrow{r_1} e_1 \xrightarrow{r_2} \dots \xrightarrow{r_l} e_l,
\end{equation*}
where $e_i\in E$, $r_i\in R$, and $l$ denotes the length of the reasoning path. For example, given the question ``\textit{Who is the brother of Justin Bieber?}'', a possible reasoning path to the answer, as shown in Figure~\ref{fig:example}, is:
\[
\small\textit{JustinBieber} \xrightarrow{\textit{has\_parent}} \textit{JeremyBieber} \xrightarrow{\textit{has\_child}} \textit{JaxonBieber}.
\]

\subsection{Benchmark Datasets}
We build our experimental datasets based on two KGQA benchmarks that are mostly used by current KG-RAG methods: WebQuestionsSP (WebQSP) \citep{yih2016value} and Complex WebQuestions (CWQ) \citep{talmor2018web}. 
These datasets consist of natural language questions designed to be answered using a KG. The questions are open-domain and require retrieval of structured information from Freebase \citep{bollacker2008freebase}, a large-scale KG with 88 million entities and 126 million triples. 
Each question in these datasets is annotated with a \emph{topic entity} and an \emph{answer entity}.
\begin{itemize}
    \item \textbf{Topic Entity}: This is the primary entity mentioned in the question, serving as the starting point for retrieving triples from the KG. In the running example, \textit{``JustinBieber''} is the topic entity.
    \item \textbf{Answer Entity}: This is the specific entity within the KG that answers the given question. In our running example, the answer entity would be \textit{``JaxonBieber''}, representing Justin Bieber's brother. Note that we might have multiple answer entities for the same question.
\end{itemize}

\subsection{KG-RAG Methods}
\textbf{RoG \citep{luo2024reasoning}} adopts a planning--retrieval--reasoning pipeline. The LLM first generates relation paths as faithful reasoning plans, which are then used to retrieve valid reasoning paths from the KG for answer generation.

\noindent\textbf{ToG \citep{sun2024think}}  treats the LLM as a graph reasoning agent. It performs iterative beam search over the KG, dynamically constructing top-ranked reasoning paths via relation/entity exploration and pruning.

\noindent\textbf{G-Retriever \citep{he2024g}} introduces a retrieval-augmented framework that extracts a compact subgraph via PCST over candidate nodes and encodes it as a soft prompt to guide LLM-based answer generation.


\subsection{Deletion Strategies}
 
To evaluate the robustness of three popular KG-RAG methods (ToG , RoG and G-Retriever) under KG incompleteness, we simulate missing knowledge by removing triples from the KG of WebQSP and CWQ. For ToG we choose  LLama2-70B-Chat as backbone model. We employ two different deletion strategies: \textbf{random triple deletion} and \textbf{reasoning path disruption}, each designed to model different levels of KG incompleteness.

\begin{table*}[t!]
\centering
\resizebox{\textwidth}{!}{
\begin{tabular}{l|ll|ll}
\toprule
\multirow{2}{*}{\textbf{Experimental Setting}} & \multicolumn{2}{c|}{\textbf{WebQSP}} & \multicolumn{2}{c}{\textbf{CWQ}}\\
& Accuracy (\%) & Hits (\%) & Accuracy (\%) & Hits (\%)\\
\midrule
RoG - w/o deletion & 76.75 & 86.60 & 57.49 & 61.79\\
RoG - 5\% random deletion & 75.55 (-1.56\%) & 85.99 (-0.71\%) & 57.33 (-0.28\%) & 61.54 (-0.40\%)\\
RoG - 10\% random deletion & 74.66 (-2.72\%) & 85.68 (-1.06\%)  & 55.88 (-2.80\%) &  60.43 (-2.20\%)\\
RoG - 20\% random deletion & 72.15 (-5.99\%) & 84.15 (-2.83\%) & 53.23 (-7.41\%) & 58.12 (-5.94\%)\\
RoG - reasoning path disruption &	65.43 (-14.70\%) &	78.12 (-9.80\%) &		52.95 (-7.90\%) & 	57.4 (-7.10\%) \\ 
RoG - w/o retrieval & 50.46 (-34.25\%) & 65.05 (-24.90\%) & 35.40 (-38.42\%) & 40.10 (-35.11\%)\\
\midrule
ToG - w/o deletion & 44.19 & 67.40 & 48.94 & 51.80\\
ToG - 5\% random deletion & 43.74 (-1.02\%)  &  67.17 (-0.33\%) & 48.78 (-0.32\%) & 51.34 (-0.89\%)\\
ToG - 10\% random deletion & 43.09 (-1.64\%) & 66.81 (-0.88\%) & 47.94 (-2.05\%) & 50.44 (-2.63\%)\\
ToG - 20\% random deletion & 41.61 (-3.42\%) &  66.18 (-1.81\%) & 46.14 (-5.73\%) & 49.18 (-5.05\%)\\
ToG - reasoning path disruption & 38.25 (-13.44\%)   & 62.70 (-6.97\%) & 44.56 (-8.95\%)   & 47.31  (-8.67\%) \\
ToG - w/o retrieval &  34.63 (-21.60\%) & 57.24 (-15.08\%) &  34.39 (-29.73\%) & 37.40 (-27.80\%)\\
\midrule
G-Retriever - w/o deletion &  53.43 & 71.49 & 35.39  & 41.51 \\
G-Retriever - 5\% random deletion & 52.33 (-2.06\%) & 70.91 (-0.81\%) & 34.63 (-2.15\%)  & 40.55 (-2.31\%) \\
G-Retriever - 10\% random deletion & 51.87 (-2.92\%) & 70.76 (-1.02\%) & 33.98 (-3.98\%)  & 40.27 (-2.99\%) \\
G-Retriever - 20\% random deletion & 50.88 (-4.78\%) & 69.70 (-2.51\%) & 33.78 (-4.55\%)  & 39.98 (-3.69\%) \\
G-Retriever - reasoning path disruption &  49.36 (-7.61\%) & 69.41 (-2.91\%) & 33.56 (-5.17\%) &  39.73 (-4.29\%)\\
G-Retriever - w/o retrieval & 29.72 (-44.37\%) & 47.17 (-34.01\%) & 14.86 (-58.01\%)  & 17.55 (-57.72\%) \\
\bottomrule
\end{tabular}}
\caption{Performance of KG-RAG methods under different levels of KG incompleteness. The numbers in parentheses indicate the relative performance drop compared to the no-deletion setting for each method.}
\label{tab:main}
\end{table*}

\paragraph{Random Triple Deletion.} 
We randomly remove a specified percentage of triples from the KG. This approach provides a baseline for understanding the general impact of missing knowledge. By varying the deletion percentage, we analyze how KG-based RAG performance degrades as more information is lost. 

\paragraph{Reasoning Path Disruption.} 
To simulate more severe knowledge incompleteness, this strategy selectively disrupts reasoning paths that support LLMs' inference. For each question, we identify the shortest reasoning paths between the topic and answer entities using the breadth-first search algorithm \citep{bfs}. We then randomly select \textbf{one} such path and remove a randomly chosen triple from it. This breaks a critical step in the reasoning chain, emulating real-world scenarios where essential intermediate knowledge may be missing.

\subsection{Evaluation Metrics}
The performance of KG-RAG is evaluated by \emph{Accuracy} and \emph{Hits}. 
Let $Q$ denote the set of test questions. 
For each question $q\in Q$, 
the KG-RAG model generates a sequence of tokens as output, denoted by $S(q)$.
Let $A(q)=\{a_1, a_2,\dots\}$ denote the corresponding set of ground truth token sequences for question $q$. Then, the evaluation metrics are defined as follows:
\begin{align*}
    Accuracy&:=\frac{1}{|Q|}\sum_{q\in Q}\sum_{a\in A(q)}\frac{\mathbbm{1}[a\preceq_{\text{subseq}}S(q)]}{|A(q)|}; \\
    \quad Hits&:=\frac{1}{|Q|}\sum_{q\in Q}\max_{a\in A(q)}\mathbbm{1}[a\preceq_{\text{subseq}}S(q)],
\end{align*}
where $\mathbbm{1}[\cdot]$ is the indicator function and $\preceq_{\text{subseq}}$ denotes the contiguous subsequence relation (i.e. $a\preceq_{\text{subseq}}S(q)$ means that the sequence $a$ appears in order within $S(q)$ without interruption).

\section{Experimental Results}
Table \ref{tab:main} presents the performance of three KG-RAG methods under varying levels of KG incompleteness. 
We evaluate models in multiple settings: a fully intact KG (no deletion), random triple deletion at 5\%, 10\%, and 20\% levels, reasoning path disruption (disabling a key step in the reasoning chain), and a baseline condition where retrieval from the KG is entirely disabled. 
Our findings highlight three key observations.

First, using KG as retrieval source significantly improves performance compared to the LLM without retrieval. For example, RoG's accuracy falls from 76.75\% to 50.46\% (-34.25\%) in WebQSP.
However, all evaluated KG-RAG methods exhibit \textbf{high sensitivity to missing knowledge in the KG}. Notably, even under random deletion, where test question-specific triples may remain unaffected, accuracy consistently declines as the proportion of deleted triples increases across both datasets. 
The relative accuracy drop is up to 5.99\% in WebQSP and 7.41\% in CWQ under 20\% random triple deletion.

Second, \textbf{disrupting a single reasoning path leads to a substantial performance drop}, despite that multiple reasoning paths often exist for a given question. 
For instance, in WebQSP, RoG's accuracy drops from 76.75\% to 65.43\% (-14.70\%), and its hits declines from 86.6\% to 78.12\% (-9.80\%).
This suggests that current KG-RAG methods often rely on specific paths for reasoning, rather than effectively utilizing alternative reasoning pathways when available. Models required fine-tuning on the KG such as RoG may be particularly dependent on a small set of learned retrieval patterns, making them vulnerable to missing knowledge in those specific paths. 

Finally, our results indicate that \textbf{KGs remain valuable retrieval sources even when incomplete}. Although performance degrades under KG incompleteness, it still surpasses the baseline where no KG retrieval is used. This suggests that addressing KG incompleteness more effectively—rather than discarding KGs entirely—should be the focus when designing more resilient KG-RAG methods. 

\section{Case Study}
We conduct a case study on the WebQTest-116 example: (\textit{``which country was Justin Bieber born in?''}). The gold reasoning path is:
\[
\small
\textit{Justin Bieber}
\xrightarrow{\textit{nationality}}
\textit{Canada}
\]

Although the question explicitly asks for a country, models often struggle when intermediate entities (e.g., cities or provinces) are involved. After removing the triple \(\langle \textit{Justin Bieber, nationality, Canada} \rangle\), all evaluated KG-RAG methods fail to recover the correct answer, despite the existence of a reasonable alternative paths:

\[
\small
\textit{Justin Bieber}
\xrightarrow{\textit{place of birth}}
\textit{London}
\xrightarrow{\textit{contained by}}
\textit{Canada}
\]
or
\[
\small
\textit{Justin Bieber}
\xrightarrow{\textit{place lived}}
\textit{Stratford}
\xrightarrow{\textit{contained by}}
\textit{Canada}
\]

A closer analysis reveals that \textsc{G-Retriever} incorrectly outputs a province (Ontario), while \textsc{RoG} retrieves a city (London), both misinterpreting the query’s focus on a country. 

To verify that this issue is not limited to a single example, we observe similar failures in more cases.
In WebQTest-436 (\textit{``What is the Nigeria time?''}), the gold reasoning path is:
\[
\small
\textit{Nigeria}
\xrightarrow{\textit{time zones}}
\textit{West Africa Time Zone}.
\]
A valid alternative path exists via an administrative region:
\[
\small
\textit{Nigeria}
\xrightarrow{\substack{\text{administrative} \\ \text{division}}}
\textit{Bauchi}
\xrightarrow{\textit{time zones}}
\textit{West Africa Time Zone}.
\]

Similarly, in WebQTest-1481 (\textit{``What city is the University of Oregon state in?''}), the gold path is:
\[
\small
\textit{University of Oregon}
\xrightarrow{\textit{contained by}}
\textit{Eugene}
\]
with a valid alternative route through the campus entity:
\[
\small
\textit{University of Oregon}
\xrightarrow{{\substack{\text{has} \\ \text{campus}}}}
\substack{\textit{Eugene}\\ \textit{Campus}}
\xrightarrow{\textit{contained by}}
\textit{Eugene}.
\]

In both cases, KG-RAG models fail to leverage the alternative paths once the gold triple is removed. This highlights a broader weakness: these models often rely on memorized or shallow retrieval patterns and struggle to generalize via multi-hop, semantically coherent reasoning chains—despite such paths being available in the KG.

\section{Conclusion and Future Work}
This work evaluated three KG-RAG methods under knowledge incompleteness, demonstrating their sensitivity to missing information. Even random triple deletions reduced performance, while reasoning path disruption led to substantial performance drops. Despite this, KGs remained valuable retrieval sources, consistently outperforming retrieval-free baselines.

Beyond incompleteness, real-world KGs also contain noise, such as incorrect triples generated during automatic KG construction process or errors in the original knowledge source \citep{heindorf2016vandalism, paulheim2016knowledge}. Additionally, the lack of a standardized evaluation protocol for KG-RAG methods and the inherent factual 
instability of LLMs  \citep{potyka2024robust, he2025supposedlyequivalentfactsarent} can lead to inconsistent assessments across studies. 

Future work should focus on developing more robust KG-RAG methods that can handle both missing and noisy knowledge, ensuring reliable performance in real-world scenarios. This includes uncertainty-aware retrieval, noise-resistant reasoning mechanisms, and hybrid approaches that integrate structured and unstructured sources. Additionally, creating comprehensive benchmarks and standardized evaluation protocols will enable systematic assessment and foster progress toward more resilient KG-RAG systems.


\section{Limitations}
Our study evaluates the robustness of KG-RAG methods to knowledge incompleteness by systematically removing triples from the source KG. While this approach reveals meaningful failure modes, one limitation is that some removed triples may not be inferable from the remaining knowledge. In such cases, performance degradation is expected and does not necessarily reflect a model’s reasoning capability. However, if such non-inferable deletions are common, this raises broader concerns about the adequacy of existing benchmarks for evaluating KG-RAG robustness.

\bibliographystyle{plainnat}
\bibliography{anthology}

\newpage

\appendix
\section{KG-RAG Method Details}
\label{appendix:kg-rag-methods}

\paragraph{RoG (Reasoning on Graphs).}
RoG implements a faithful reasoning pipeline by decomposing KGQA into three stages: 
(1) \textit{planning}: prompting the LLM to generate relation paths grounded in the KG (e.g., \texttt{marry\_to → father\_of}); 
(2) \textit{retrieval}: matching these relation paths with actual reasoning paths in the KG (e.g., \texttt{Alice → Bob → Charlie}); 
(3) \textit{reasoning}: using the retrieved paths as evidence to answer the question.

Formally, RoG models the answer probability as:
\begin{equation}
P_\theta(a \mid q, G) = \sum_{z \in \mathcal{Z}} P_\theta(a \mid q, z, G) \cdot P_\theta(z \mid q)
\end{equation}
where \( z \) denotes a relation path (plan), and \( \theta \) are the LLM parameters.

Training is guided by the ELBO \citep{DBLP:journals/corr/KingmaW13}:
\begin{equation}
\log P(a \mid q, G)\geq\mathbb{E}_{z \sim Q(z)} \big[ \log P_\theta(a \mid q, z, G) \big] - \mathrm{KL}\left( Q(z) \,\Vert\, P_\theta(z \mid q) \right)
\end{equation}

Note that this ELBO formulation serves as a theoretical framework; in practice, RoG adopts supervised instruction tuning rather than explicitly optimizing this bound.

\paragraph{ToG (Think-on-Graph).}
ToG follows a tight integration paradigm (denoted as ``LLM$\otimes$KG'' in the original paper), where LLMs interact directly with the KG during reasoning via iterative exploration and pruning.
It consists of three phases:

\begin{itemize}
  \item \textbf{Initialization}: The LLM identifies top-N topic entities from the question.
  \item \textbf{Exploration}: At each iteration, the LLM conducts relation and entity exploration from current frontiers and prunes candidates to retain top-N reasoning paths.
  \item \textbf{Reasoning}: The LLM evaluates whether the current paths suffice to answer the question. If not, exploration continues until a satisfactory path is found or a max depth is reached.
\end{itemize}

\paragraph{G-Retriever.}
G-Retriever is a retrieval-augmented generation framework designed for question answering over graphs. It is applicable to both structured symbolic graphs (e.g., Freebase) and text-attributed graphs (e.g., scene graphs, commonsense graphs). 

The method integrates subgraph retrieval with prompt-based conditioning as follows:
\begin{itemize}
\item \textbf{Subgraph Retrieval via PCST}: Given a question and topic entity, G-Retriever first selects candidate nodes by exploring the local KG neighborhood (e.g., 1–2 hops). These nodes are treated as rewards in a Prize-Collecting Steiner Tree (PCST) formulation, which extracts a compact subgraph containing informative multi-hop paths.

  \item \textbf{Graph Prompt Construction}: The retrieved subgraph is linearized into token sequences using fixed templates (e.g., \texttt{[head] --[relation]--> [tail]}), converting structural triples into flattened sequences.

  \item \textbf{Soft Prompt Encoding}: Rather than appending text directly, the serialized subgraph is embedded as a trainable soft prompt. These embeddings are prepended to the LLM input, conditioning generation in a parameter-efficient manner.
\end{itemize}

At inference time, the LLM processes the soft prompt and question jointly to generate the answer. This design enables faithful reasoning over retrieved evidence while adapting to different graph modalities through prompt tuning.

\end{document}